\documentclass[11pt]{article}

\usepackage[final]{acl}

\usepackage{times}
\usepackage{latexsym}
\usepackage{amsmath}
\usepackage{amsfonts}
\usepackage{amssymb}
\usepackage{multirow}
\providecommand{\mathbb}[1]{\mathbf{#1}}
\usepackage{authblk}

\usepackage[T1]{fontenc}

\usepackage[utf8]{inputenc}

\usepackage{microtype}

\usepackage{inconsolata}

\usepackage{graphicx}
\usepackage{booktabs}
\usepackage{algorithm}
\usepackage{algpseudocode}
\usepackage[table]{xcolor}
\usepackage{makecell}
\usepackage{adjustbox}
\usepackage{array}

\definecolor{grayA}{HTML}{F7F7F7}
\definecolor{grayB}{HTML}{EFEFEF}
\definecolor{grayC}{HTML}{E6E6E6}
\definecolor{oursblue}{HTML}{EAF7FB}

\title{EMS: Multi-Agent Voting via Efficient Majority-then-Stopping}

\author[1]{Yiqing Liu}
\author[1]{Hantao Yao}
\author[1]{Wu Liu}
\author[1]{Yongdong Zhang}
\affil[1]{University of Science and Technology of China}

\begin{document}
{\makeatletter\acl@finalcopytrue
\maketitle}
\begin{abstract}

Majority voting is the standard for aggregating multi-agent responses into a final decision.
However, traditional methods typically require all agents to complete their reasoning before aggregation begins, leading to significant computational overhead, as many responses become redundant once a majority consensus is achieved. 
In this work, we formulate efficient multi-agent voting as a reliability-aware agent scheduling problem and propose Efficient Majority-then-Stopping (EMS) to improve reasoning efficiency. 
EMS first estimates a Task-Conditioned Reliability Ordering (TCRO) for each agent by retrieving its historical consensus evidence on semantically similar queries, and then invoking agents in descending reliability order. 
Next, Adaptive Incremental Voting (AIV) terminates the process once the current leading answer cannot be overturned by any possible votes from the remaining agents, and returns this answer.
Finally, Reliability History Updating (RHU) updates only the invoked agents according to their consensus with the final decision. 
Extensive evaluations across five benchmarks show that EMS preserves the accuracy of Majority Voting while reducing the average number of invoked agents by $35\%$ and  token consumption by $44\%$, respectively.
The code is available at \url{https://github.com/fuyu66/EMS}.

\end{abstract}

\section{Introduction}

Large Language Models (LLMs) have demonstrated remarkable success in complex tasks such as mathematical reasoning, code generation, and commonsense questions answering~\cite{kojima2022large,wan2023better,li2025fundamental}. 
Despite these advancements, a single LLM often exhibits limited reasoning diversity and struggles with problems requiring dynamic, task-specific strategies~\cite{mirzadeh2024gsm,jiang2024peek,guo2024large}. 
Recent research has sought to address these limitations by leveraging multi-agent systems (MAS)~\cite{li2023theory,chen2023agentverse,wu2024autogen}, where multiple LLM-based agents collaborate to tackle complex tasks.
However, recent studies suggest that the gains of MAS are not necessarily driven by interaction-heavy communication. 
\citet{choi2025debate,zhou2025multi} show that aggregation can be regarded as a fundamental topology in MAS and can achieve performance comparable to debate in reasoning workflows.

\begin{figure}
	\centering
	\includegraphics[width=1\linewidth,height=0.75\linewidth]{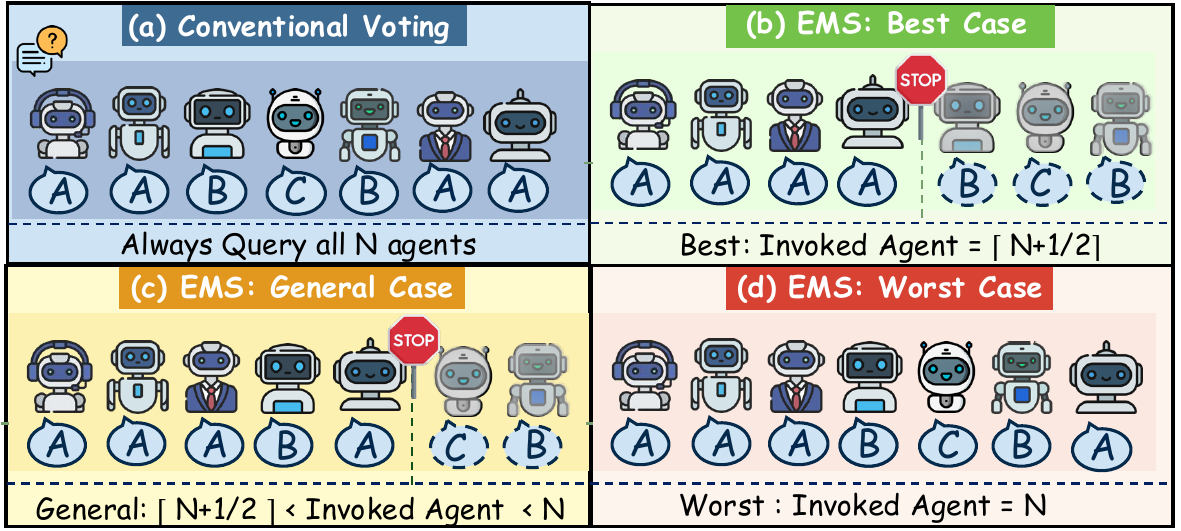}
    \caption{Intuitive comparison of different multi-agent voting processes. There exists a significant difference in terms of \emph{Number of Invoked Agents} among different voting sequences.}
    \label{fig:f1}
\end{figure}

Among the various aggregation techniques, majority voting has become one of the most widely used methods due to its simplicity and effectiveness\cite{DBLP:conf/emnlp/ZhaoWP24,kaesberg2025voting}. 
In the standard majority voting setting, each agent performs reasoning independently, and the final decision is made based on the most frequent response. 
Several extensions of this basic paradigm have been proposed to enhance performance, such as reliability-based weighting considering agents' past accuracy, and confidence-based soft voting integrating model-reported confidence scores. 
However, these approaches still follow the same fundamental paradigm: \emph{reasoning first, aggregation later}. 
In this setup, all agents must complete their reasoning before a final decision can be made. 
In many scenarios, the majority decision can be reached before all agents have completed their reasoning, making the contributions of the remaining agents redundant. 
This leads to unnecessary computational overhead, especially when large numbers of agents or expensive LLM inferences are involved. 
This inefficiency raises a key challenge: \emph{How can we reduce redundant reasoning in multi-agent LLM systems while preserving the accuracy benefits of majority voting?}

A natural insight to improve the efficiency of majority voting is to reduce reasoning that can no longer affect the final decision. 
In a system with $N$ agents, once the current leading answer is guaranteed to remain the final winner regardless of how the remaining agents vote, the system can safely terminate and return the certified answer. 
This observation suggests a majority-then-stop mechanism that preserves the decision rule of standard majority voting while reducing redundant inference. 
However, the efficiency of such early stopping critically depends on the order in which agents are invoked. 
As shown in Figure~\ref{fig:f1}, if the most reliable agents are consulted first, a consensus can be reached more quickly, whereas querying less reliable agents early may require waiting for more agents to vote.
Therefore, achieving efficient early stopping requires prioritizing agents that are most likely to produce a consensus for majority voting.

To address this issue, we formulate multi-agent majority voting as a reliability-aware agent scheduling problem, 
and propose Efficient Majority-then-Stopping (EMS) to reduce inference cost while preserving the decision quality of full majority voting. 
EMS estimates a task-conditioned reliability score for each agent based on its historical agreement evidence retrieved according to the semantic similarity of the current query. 
The agents are then invoked in descending order of this score, allowing the system to reach a certified majority decision as early as possible. 
Specifically, EMS consists of three steps. First, Task-Conditioned Reliability Ordering (TCRO) ranks agents according to their expected reliability on the current query. 
Second, Adaptive Incremental Voting (AIV) invokes agents following this order and terminates the voting process once the current leading answer is certified to be the final winner. 
Third, Reliability History Updating (RHU) records the consensus of each invoked agent and updates its local history. 


\begin{enumerate}
\item We analyze the inefficiency of the widely used reasoning-first aggregation paradigm in Multi-Agent Voting and formulate it as a reliability-aware agent scheduling problem to improve inference efficiency.
\item We propose an Efficient Majority-then-Stopping (EMS), which combines task-conditioned  reliability ordering with adaptive incremental voting to terminate redundant agent invocations safely.
\item  Experiments on five benchmarks show that
EMS  reduces ``Avg.\#Agents'' and ``Avg.Tokens'' by 35\% and 44\% while preserving the accuracy of Majority Voting, respectively.
\end{enumerate}

\section{Related Work}

\paragraph{Multi-Agent Systems.}
Recent advances in large language models (LLMs) have stimulated increasing interest in multi-agent systems (MAS)~\cite{kojima2022large,wan2023better,mirzadeh2024gsm}, where multiple LLM-based agents collaborate to solve complex tasks. 
Representative frameworks such as AutoGen \cite{wu2024autogen} enable structured collaboration among multiple agents through role assignment, communication, and task decomposition.
Other interaction-based approaches, including multi-agent debate~\cite{chan2023chateval,liang2024encouraging}, further encourage agents to critique and refine one another's intermediate reasoning, thereby improving the quality of final outputs. 
While these studies demonstrate the effectiveness of collaborative reasoning for improving task performance, the efficiency of multi-agent inference, particularly the reduction of redundant agent calls during decision making, remains relatively underexplored.

\begin{figure*}
    \centering
    \includegraphics[width=1\linewidth, height=0.45\linewidth]{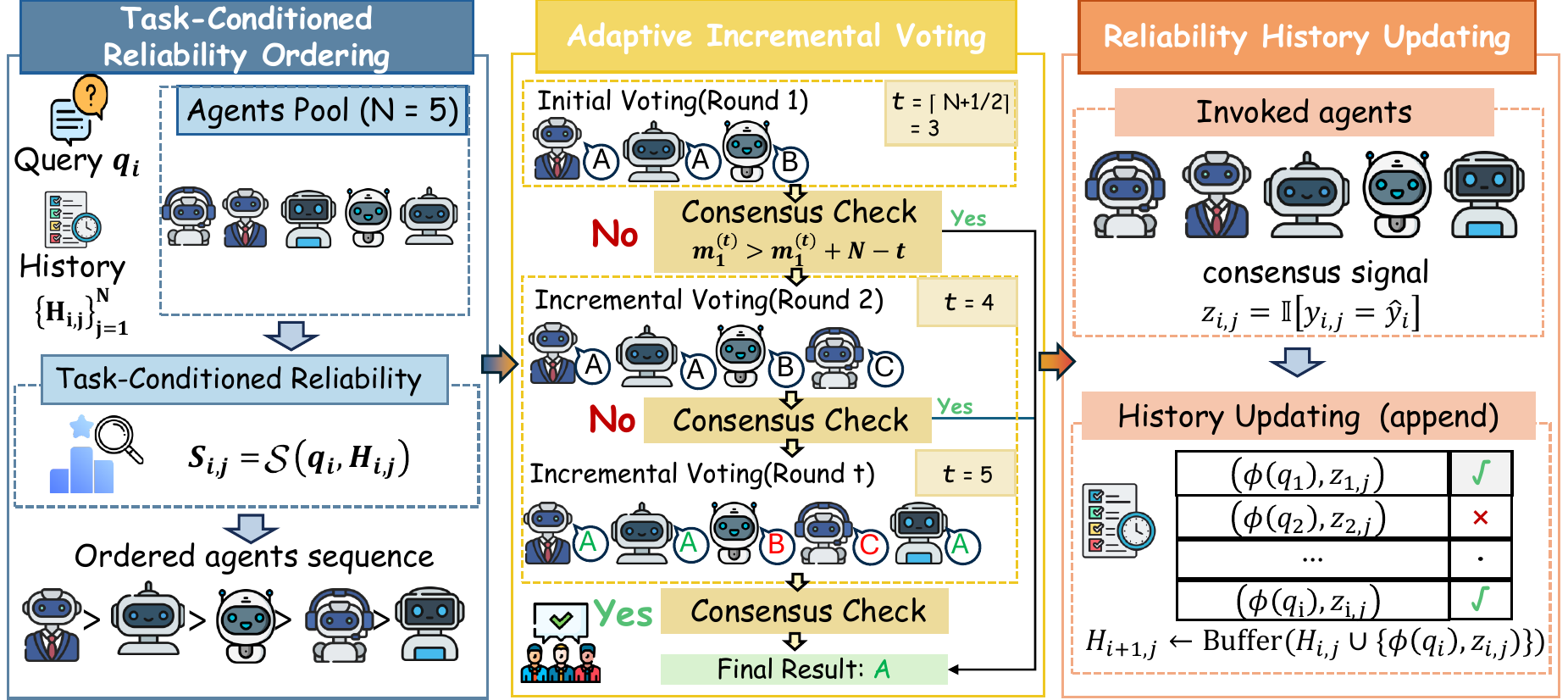}
    \caption{Overview of the proposed Efficient Majority-then-Stopping (EMS) framework. For each query, EMS first uses the Task-Conditioned Reliability Ordering (TCRO) to determine a reliability-aware voting order. It then performs incremental voting via Adaptive Incremental Voting (AIV), where agents are invoked sequentially according to the estimated order. Finally, Reliability History Updating (RHU) updates the confidence state of the agents contributing to the majority-voting.}
    \label{fig:f2}
\end{figure*}

\paragraph{Efficient Multi-Agent Inference.} 
To improve efficiency, recent work has explored adaptive routing and selective invocation mechanisms \cite{yue2025masrouter}.
For example, MASRouter~\cite{yue2025masrouter} learns to route LLMs in
multi-agent systems, reducing unnecessary model calls by selectively assigning
agents to different inputs. DOWN~\cite{eo2025debate}activates debate only when initial confidence suggests that additional deliberation is necessary, thereby avoiding unnecessary interaction on easy instances. 
Other methods, such as AgentDropout~\cite{DBLP:conf/acl/WangW00Z0025} and
ARG-Designer~\cite{DBLP:conf/aaai/LiLWZP26}, further improve collaborative reasoning by pruning redundant agents or optimizing interaction structures. These approaches primarily reduce the cost of debate or collaboration, whereas EMS directly optimizes the aggregation stage by exploiting the majority-voting structure for certified early termination.

\paragraph{Multi-Agent Voting and Aggregation.}
Aggregating outputs from multiple agents is a fundamental component of MAS \cite{DBLP:conf/coling/WangLF0W25}, and majority voting remains one of the most widely adopted decision rules because of its simplicity and strong performance.
Recent work demonstrates that majority voting accounts for the vast majority of performance gains typically attributed to complex multi-agent debate, suggesting that expensive inter-agent communication rounds are often unnecessary \cite{kaesberg2025voting,choi2025debate}. Furthermore, standard voting protocols are particularly optimal for reasoning-based tasks, significantly outperforming other decision-making structures \cite{DBLP:journals/corr/abs-2511-11040}. 
Despite these advances, most existing approaches follow a \emph{reasoning-first-aggregation-later} 
paradigm, which introduces significant computational waste as the final decision is often reachable before all agents complete their reasoning. Our work addresses this gap by formulating majority voting as a reliability-aware agent scheduling problem.

\section{Methodology}
\subsection{Problem Formulation}

Consider a multi-agent system~\cite{li2023theory,kim2025towards} composed of
\(N\) agents
\(\mathcal{A}=\{\boldsymbol{a}_1,\boldsymbol{a}_2,\ldots,\boldsymbol{a}_N\}\),
where each agent \(\boldsymbol{a}_j\) is instantiated by a specific Large
Language Model (LLM). Let
\(\mathcal{Q}=\{\boldsymbol{q}_1,\boldsymbol{q}_2,\ldots,\boldsymbol{q}_M\}\)
denote a dataset consisting of \(M\) queries, where each query
\(\boldsymbol{q}_i\) is associated with a ground-truth answer \(y_i^\star\).
The ground-truth answer is used only for evaluation and is not accessible
during inference.

Each agent \(\boldsymbol{a}_j\) is modeled as an inference function
\(\mathcal{F}_j:\mathcal{Q}\rightarrow\mathcal{Y}\), where \(\mathcal{Y}\)
denotes the answer space. Given a query \(\boldsymbol{q}_i\), agent
\(\boldsymbol{a}_j\) independently produces a prediction
\(
y_{i,j}=\mathcal{F}_j(\boldsymbol{q}_i),
\)
where the second subscript \(j\) always refers to the original index of the
agent in \(\mathcal{A}\). The goal of multi-agent decision making is to
produce a final prediction \(\hat{y}_i\) per query by aggregating the
responses of invoked agents.

\emph{Vanilla Majority-Voting.}
In the vanilla majority-voting scheme, all \(N\) agents are invoked for each
query \(\boldsymbol{q}_i\). The full response set is
\begin{equation}
Y_i^{\mathrm{full}}
=
\{y_{i,1},y_{i,2},\ldots,y_{i,N}\}.
\end{equation}
Let \(\mathcal{U}_i^{\mathrm{full}}\) denote the set of unique answers
appearing in \(Y_i^{\mathrm{full}}\). The final prediction is given by
\begin{equation}
\label{eq:majority_vote}
\hat{y}_i^{\mathrm{MV}}
=
\arg\max_{y\in\mathcal{U}_i^{\mathrm{full}}}
\sum_{j=1}^{N}\mathbb{I}\big[y_{i,j}=y\big],
\end{equation}
where \(\mathbb{I}[\cdot]\) is the indicator function. Although the majority voting is simple and effective, it requires every agent to perform inference before
aggregation, leading to unnecessary computational cost when the final decision
could already be determined from a partial set of votes.

\emph{Sequential Majority Voting.}
To reduce redundant  invocations, we reformulate the voting process as a
sequential decision procedure. For each query \(\boldsymbol{q}_i\), the system
estimates a reliability score for each agent
\(\boldsymbol{a}_j\):
\begin{equation}
S_{i,j}
=
\mathcal{S}(\boldsymbol{q}_i, \mathcal{H}_{i,j}),
\label{eq:ordering_score}
\end{equation}
where \(\mathcal{H}_{i,j}\) denotes the history buffer maintained for
\(\boldsymbol{a}_j\) before processing \(\boldsymbol{q}_i\), and \(\mathcal{S}(\cdot,\cdot)\) denotes the reliability scoring function.
Sorting all agents by descending scores yields a permutation
\begin{equation}
\Psi_i
=
\mathrm{argsort}_{j\in\{1,\ldots,N\}}^{\downarrow}
\left(S_{i,j}\right),
\label{eq:agent_order}
\end{equation}
where \(\Psi_i(k)\) denotes the original index of the agent ranked at the
\(k\)-th position. The corresponding prioritized agent sequence is
\begin{equation}
\begin{split}
\hat{\mathcal{A}}_i
&=
(\hat{\boldsymbol{a}}_{i,1},
 \hat{\boldsymbol{a}}_{i,2},
 \ldots,
 \hat{\boldsymbol{a}}_{i,N}) \\
&=
(\boldsymbol{a}_{\Psi_i(1)},
 \boldsymbol{a}_{\Psi_i(2)},
 \ldots,
 \boldsymbol{a}_{\Psi_i(N)}).
\label{eq:ordered_agents}
\end{split}
\end{equation}

After the \(t\) agents in the prioritized sequence have been invoked,
the set of original indices is
\begin{equation}
\mathcal{I}_i(t)
=
\{\Psi_i(1),\Psi_i(2),\ldots,\Psi_i(t)\}.
\end{equation}
The corresponding partial response set is
\begin{equation}
\begin{aligned}
Y_i(t)
&=
\{y_{i,j}\mid j\in\mathcal{I}_i(t)\} \\
&=
\{y_{i,\Psi_i(1)},y_{i,\Psi_i(2)},\ldots,y_{i,\Psi_i(t)}\}.
\end{aligned}
\end{equation}
Here, \(Y_i(t)\) contains only the responses of the agents that have actually
been invoked by step \(t\), rather than the full response set
\(Y_i^{\mathrm{full}}\).

Let \(L_i\leq N\) denote the stopping time of the sequential voting process,
i.e., the number of agents invoked before the system terminates. The final
prediction \(\hat{y}_i\) is computed from the partial response set
\(Y_i(L_i)\). The accuracy of \(\hat{y}_i\) is evaluated by comparing it with
the ground-truth answer \(y_i^\star\).

The objective is to preserve the accuracy of full majority-voting while
reducing the expected number of invoked agents:
\begin{equation}
\max_{\Pi}
\;
\mathbb{E}_{\boldsymbol{q}_i\sim\mathcal{Q}}
\left[
\mathbb{I}[\hat{y}_i=y_i^\star]
\right]
-
\lambda
\mathbb{E}_{\boldsymbol{q}_i\sim\mathcal{Q}}
\left[
\frac{L_i}{N}
\right],
\label{eq:objective}
\end{equation}
where \(\Pi\) denotes the sequential voting policy, including agent ordering
and stopping, and \(\lambda\geq 0\) controls the accuracy-cost trade-off.

The main insight of EMS is that it treats the multi-agent voting process as a sequential problem, where agents vote one after another. As illustrated in Figure~\ref{fig:f2}, EMS consists of three main stages: 

\subsection{Task-Conditioned Reliability Ordering}
The key to efficient sequential voting is to invoke more reliable agents earlier.
Since heterogeneous LLM agents may show
different strengths across tasks, 
we propose a task-conditioned
reliability score for each agent, dynamically estimate the reliability of each agent conditioned on the explicit semantic context of the current query.


For each agent \(\boldsymbol{a}_j\), we define its task-conditioned reliability score on query
\(\boldsymbol{q}_i\) as a smoothed posterior estimate:
\begin{equation}
S_{i,j}
=
\frac{
\rho+m_{i,j}\,\bar{z}_{i,j}^{(k)}
}{
2\rho+m_{i,j}
},
\label{eq:tc_score}
\end{equation}
where \(m_{i,j}\) denotes the amount of retrieved local evidence, \(\bar{z}_{i,j}^{(k)}\) is the top-\(k\) task-conditioned agreement
rate, and
\(\rho>0\) is the strength of a symmetric Beta prior. This prior shrinks the
estimate toward the neutral value \(1/2\) when only limited historical evidence
is available.
We next define the quantities in Eq.~\eqref{eq:tc_score}. Before processing
query \(\boldsymbol{q}_i\), agent \(\boldsymbol{a}_j\) maintains a growing history
buffer \(\mathcal{H}_{i,j}\). For each previous query \(\boldsymbol{q}_h\) on
which \(\boldsymbol{a}_j\) was invoked, we store an evidence tuple
\(
(e_h,z_{h,j})\in\mathcal{H}_{i,j},
\)
where \(e_h=\phi(\boldsymbol{q}_h)\) is the query embedding produced by a
sentence encoder \(\phi(\cdot)\). And
\begin{equation}
z_{h,j}
=
\mathbb{I}[y_{h,j}=\hat{y}_h]
\label{eq:agreement_indicator}
\end{equation}
indicates whether the answer of \(\boldsymbol{a}_j\) agreed with the final
voting consensus \(\hat{y}_h\). 

Given the current query \(\boldsymbol{q}_i\), let
\(e_i=\phi(\boldsymbol{q}_i)\). For agent \(\boldsymbol{a}_j\), we retrieve the
top-\(k\) historical evidence tuples that are most similar to the current
query:
\begin{equation}
\mathcal{N}_{i,j}^{(k)}
=
\operatorname{TopK}_{(e_h,z_{h,j})\in\mathcal{H}_{i,j}}
\left(\cos(e_i,e_h)\right),
\label{eq:topk_set}
\end{equation}
where \(\cos(e_i,e_h)\) denotes the cosine similarity between the current
query and a historical query. 
If the history buffer is empty, then \(\mathcal{N}_{i,j}^{(k)}=\varnothing\).
Then the amount of retrieved local evidence is
\begin{equation}
m_{i,j}
=
|\mathcal{N}_{i,j}^{(k)}|,
\label{eq:evidence_size}
\end{equation}
where \(|\cdot|\) denotes set cardinality. 

For each retrieved tuple
\((e_h,z_{h,j})\in\mathcal{N}_{i,j}^{(k)}\), we assign a top-\(k\) softmax
weight:
\begin{equation}
\omega_{i,h}^{(j)}
=
\frac{
\exp(\cos(e_i,e_h))
}{
\sum_{(e_{h'},z_{h',j})\in\mathcal{N}_{i,j}^{(k)}}
\exp(\cos(e_i,e_{h'}))
}.
\label{eq:topk_softmax_weight}
\end{equation}
Here, \(h'\) is a dummy index over the retrieved evidence tuples. The softmax
assigns larger weights to historical queries that are more semantically
similar to the current query.
The top-\(k\) task-conditioned agreement rate is then
\begin{equation}
\bar{z}_{i,j}^{(k)}
=
\sum_{(e_h,z_{h,j})\in\mathcal{N}_{i,j}^{(k)}}
\omega_{i,h}^{(j)}\,\cdot z_{h,j},
\label{eq:weighted_agreement}
\end{equation}

When no historical evidence is available, \(m_{i,j}=0\), and
Eq.~\eqref{eq:tc_score} reduces to \(S_{i,j}=1/2\), so all agents start from a
neutral reliability estimate. 
After computing \(S_{i,j}\) for all
agents, we sort the agents in descending order according to
Eq.~\eqref{eq:agent_order} and obtain the prioritized sequence in
Eq.~\eqref{eq:ordered_agents}.

\subsection{Adaptive Incremental Voting} 

Given the ordered agent sequence
\(\widehat{\mathcal{A}}_i=(\hat{\boldsymbol{a}}_{i,1},\hat{\boldsymbol{a}}_{i,2},\ldots,\hat{\boldsymbol{a}}_{i,N})\) for query $\boldsymbol{q}_i$,
we perform adaptive incremental voting to reduce redundant reasoning. We define the initial quorum size as the smallest number of votes that can
possibly certify a final winner under the strongest initial agreement. If
fewer than half of the agents are invoked, even a unanimous partial vote can
still be overturned by the remaining agents. Therefore, the initial quorum is
set to
\(
\tau=\left\lceil\frac{N+1}{2}\right\rceil.
\)

For query $\boldsymbol{q}_i$, the first \(\tau\) agents in the ordered sequence are
invoked in parallel:
\(
\mathcal{I}_i(\tau)=\{\Psi_i(1),\ldots,\Psi_i(\tau)\}.
\)
And the partial response set is
\(
Y_i(\tau)=\{y_{i,j}\mid j\in\mathcal{I}_i(\tau)\}.
\)
If the current vote distribution certifies a final decision, the system stops.
Otherwise, the next agent in the ordered sequence is invoked, and the procedure
continues until either a certificate is obtained or all \(N\) agents have been
used.

To define the stopping rule, let \(n_i(y,t)\) be the number of votes received
by answer \(y\) after \(t\) agents have been invoked:
\begin{equation}
n_i(y,t)
=
\sum_{j\in\mathcal{I}_i(t)}
\mathbb{I}[y_{i,j}=y].
\label{eq:vote_count}
\end{equation}
Let \(m_i^{(1)}(t)\) and \(m_i^{(2)}(t)\) denote the largest and second largest
vote counts among all answers observed in \(Y_i(t)\), respectively:
\begin{equation}
\begin{aligned}
m_i^{(1)}(t)
&= \max_{y\in\mathcal{U}_i(t)} n_i(y,t), \\[4pt]
m_i^{(2)}(t)
&= \max_{y\in\mathcal{U}_i(t)\setminus\{y_i^{(1)}(t)\}} n_i(y,t),
\end{aligned}
\label{eq:your_label}
\end{equation}
where \(\mathcal{U}_i(t)\) is the set of unique answers in \(Y_i(t)\), and
\(y_i^{(1)}(t)\) is the current leading answer. If there is only one unique
answer, we set \(m_i^{(2)}(t)=0\). The number of remaining uninvoked agents is \(N-t\).

\paragraph{Theorem 1: Certified Plurality Stopping.}
After \(t\) agents have voted, if the current vote counts satisfy
\begin{equation}
m_i^{(1)}(t)>m_i^{(2)}(t)+N-t,
\label{eq:certified_plurality}
\end{equation}
then the current leading answer \(y_i^{(1)}(t)\) is guaranteed to remain the
final plurality winner regardless of how all remaining agents vote.

When Eq.~\eqref{eq:certified_plurality} holds, the voting process terminates
and returns he final prediction
\begin{equation}
\hat{y}_i=y_i^{(1)}(t).
\label{eq:early_return}
\end{equation}
Otherwise, the next agent
\(\hat{\boldsymbol{a}}_{i,t+1}\) is invoked:
\begin{equation}
\mathcal{I}_i(t+1)
=
\mathcal{I}_i(t)\cup\{\Psi_i(t+1)\},
\label{eq:incremental_agent_set}
\end{equation}
and its response is added to the partial response set:
\begin{equation}
Y_i(t+1)
=
Y_i(t)\cup\{y_{i,\Psi_i(t+1)}\}.
\label{eq:incremental_answer_set}
\end{equation}
The certificate in Eq.~\eqref{eq:certified_plurality} is then checked again.
If no certificate is obtained before all agents are invoked, the method reduces
to vanilla majority-voting over the full response set.

The certified plurality criterion 
explicitly accounts for the current runner-up and the number of remaining agents, thereby
allowing the system to stop exactly. 

\subsection{Reliability History Updating} 

After the final answer \(\hat{y}_i\) is obtained for query $\boldsymbol{q}_i$, the system
updates only the states of the agents that were actually invoked. Let
\(\mathcal{I}_i=\mathcal{I}_i(L_i)\) denote the final invoked agent set for
query $\boldsymbol{q}_i$, where \(L_i\leq N\). For each invoked agent $\boldsymbol{a}_j$ with
\(j\in\mathcal{I}_i\), we define the agreement signal
\(
z_{i,j}
=
\mathbb{I}[y_{i,j}=\hat{y}_i].
\)
The online state is updated as
\begin{equation}
\begin{aligned}
\mathcal{H}_{i+1,j} &= \operatorname{Buffer}\bigl( \mathcal{H}_{i,j} \cup \{(e_i, z_{i,j})\} \bigr).
\end{aligned}
\label{eq:online_update}
\end{equation}
where \(e_i=\phi(q_i)\), and \(\mathrm{Buffer}(\cdot)\) denotes a bounded
sliding-window History operation that retains the most recent evidence tuples
when the buffer exceeds its capacity. For agents that are not invoked,
\(j\notin\mathcal{I}_i\), the state remains unchanged.

\definecolor{grayA}{HTML}{F7F7F7}
\definecolor{grayB}{HTML}{EFEFEF}
\definecolor{grayC}{HTML}{E6E6E6}
\definecolor{oursblue}{HTML}{EAF7FB}

\begin{table*}[t]
\centering
\fontsize{10pt}{10pt}\selectfont
\setlength{\tabcolsep}{3pt}
\renewcommand{\arraystretch}{1.1}

\caption{Comparison on five benchmarks. Accuracy (\%) and efficiency metrics are reported.  
``Avg. \#Agents'' denotes the average number of agents invoked per query, and 
``Avg. Tokens'' denotes token consumption per query.}
\label{tab:main_results}

\begin{adjustbox}{max width=\textwidth}
\begin{tabular}{l|cccccc|cc}
\toprule
\multirow{2}{*}{\textbf{Methods}}
& \multicolumn{6}{c|}{\textbf{Accuracy (\%)}} 
& \multicolumn{2}{c}{\textbf{Efficiency}} \\
\cmidrule(lr){2-7} \cmidrule(lr){8-9}
& \textbf{GSM8K}
& \textbf{GPQA}
& \textbf{CSQA}
& \textbf{AQuA}
& \textbf{MMLU}
& \textbf{Avg.}
& \textbf{Avg. \#Agents}
& \textbf{Avg. Tokens} \\
\midrule

\rowcolor{grayA}
Single Agent
& 95.68  &  66.16 & 84.66 & 88.52 & 88.81
& 84.77 & 1.00 & 468 \\

\rowcolor{grayA}
Self-Consistency
& 96.73  & 66.67 & 85.88 & 88.98 & 90.63
& 85.78 & 9.00 & 4,357 \\

\midrule

\rowcolor{grayB}
MAD$^{*}$
& 96.96  & 69.17 & 85.24 & 89.92 & 89.86
& 86.23 & 6.00 &  8,499\\

\rowcolor{grayB}
AgentDropout$^{*}$

& 97.04  & 68.69 & 87.47 & 90.16 & 90.09
& 86.69 & 6.00 & 7,258 \\
\rowcolor{grayB}

\midrule

\rowcolor{grayC}
Simple MV
& 97.27  & 70.20 & 86.24 & 90.94 & 90.37
& 87.01 & 9.00 & 7,381 \\

\rowcolor{grayC}
Weighted MV
& 97.27  & 71.72 & 86.65 & 90.55 & 90.47
& 87.33 & 9.00 & 7,381 \\

\rowcolor{oursblue}
\textbf{EMS (Ours)}
& 97.27  & 70.20 & 86.24 & 90.94 & 90.50
& 87.01 & 5.84 & 4,123 \\

\bottomrule
\end{tabular}
\end{adjustbox}
\end{table*}

\section{Experiments}

\subsection{Experimental Setup}
\textbf{Datasets.}
We evaluate the performance of the proposed EMS across five benchmarks:
1) Mathematical Reasoning: 
\emph{AQuA}~\cite{ling2017program},
and 
\emph{GSM8K}~\cite{cobbe2021training}.
2) General Knowledge: \emph{MMLU}~\cite{hendrycksmeasuring}, \emph{GPQA Diamond}~\cite{rein2024gpqa}, \emph{CommonsenseQA}~\cite{talmor2019commonsenseqa}. 
All datasets utilize standard settings, with details provided in Appendix.

\noindent\textbf{Agent Configuration.}
Our framework utilizes a heterogeneous pool of $N=9$ LLMs to ensure diverse reasoning perspectives, which includes: 
1) \emph{OpenAI} (GPT-4.1, GPT-4.1-mini, GPT-5-mini);
2) \emph{Google} (Gemini-2.0-Flash);
3) \emph{Anthropic} (Claude-Haiku-4.5);
4) \emph{DeepSeek} (DeepSeek-V3);
5) \emph{Alibaba} (Qwen3-235B-A22B);
6) \emph{Meta} (Llama-4-Maverick);
and 7) \emph{Mistral AI} (Mistral-Medium-3.5);
All model inferences are conducted via the API aggregation service provided by 
Yunwu and OpenRouter.
For semantic query feature extraction, we employ the \emph{paraphrase-multilingual-MiniLM-L12-v2} encoder to map queries into a shared latent space for similarity-based retrieval.

\noindent\textbf{Baselines.}
We compare EMS against the following baselines:
1) \emph{Individual Baselines} include Single Agent with Chain-of-Thought (CoT)~\cite{DBLP:conf/nips/Wei0SBIXCLZ22}  using the top-performing model and Self-Consistency (SC)~\cite{wangself} via repeated sampling. For a fair comparison under a similar agent-calls budget, MAD uses the top-3 models with two debate rounds. For AgentDropout, we follow its heterogeneous graph setting with seven candidate agents and one communication round, and apply node dropout at inference by removing one low-importance agent per round. This gives an expected budget of six LLM calls per question, which is comparable to the baselines and to EMS.
3) \emph{Voting Baselines} include the Simple Majority Voting (Simple MV) using the full agent pool, and {Weighting methods} comprise  Historical-based Weighted MV\cite{DBLP:journals/tmlr/LiZY0Y24}.

\noindent\textbf{Evaluation Metrics.} 
Average Accuracy measures final consensus correctness, denoted as ``\emph{Avg.Acc.}''. 
Moreover, we measure multi-agent system efficiency by the \emph{Average Number of Invoked Agents}, defined as the mean number of agents called to produce responses per query. 
Formally, given a query set $\mathcal{Q}$, the metric is computed as
\begin{equation}
\text{Avg.\#Agents} = \frac{1}{|\mathcal{Q}|} \sum_{q_i \in \mathcal{Q}} \mathcal{L}(\boldsymbol{q}_i),
\end{equation}
where $\mathcal{L}(\boldsymbol{q}_i)$ denotes the number of agents invoked for query $\boldsymbol{q}_i$. 
This metric serves as a proxy for inference cost, as each agent invocation typically corresponds to one LLM call.
Average Token Consumption, denoted as ``Avg. Tokens'', measures the average token consumption per query.

\begin{figure*}[t]
	\centering
	\includegraphics[width=1\linewidth, height=0.2\linewidth]{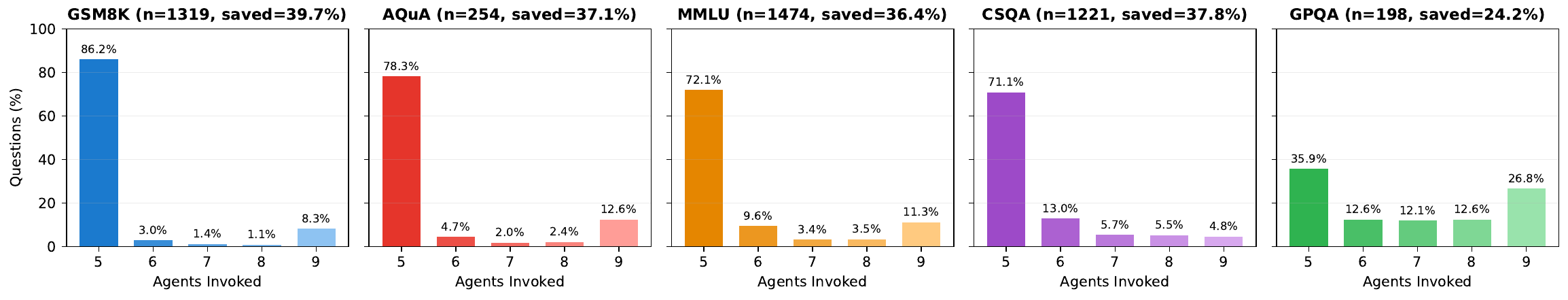}
    \caption{Analysis of the Adaptive Incremental Voting. Each bar represents the proportion of questions that achieve the final results based on the ``Agents Invoked''.}
 \label{fig:1}
\end{figure*}

\subsection{Main Results}\label{sec:main_results}
Table~\ref{tab:main_results} presents the overall comparison between EMS and existing baselines. 1) Multi-Agent  methods consistently improve over single-agent inference. 
Specifically, EMS achieves an average accuracy of 87.01\%, 
outperforming the Single Agent and Self-Consistency baselines by 2.24 and 1.23 percentage points, respectively. 
This confirms the effectiveness of aggregating multiple agents for robust reasoning.
2) EMS also compares favorably with existing Multi-Agent Collaboration baselines. It improves over MAD by 0.78 percentage points in average accuracy while invoking slightly fewer agents on average. 
Compared with efficiency-oriented methods, EMS achieves comparable or slightly better accuracy.
3) EMS achieves comparable accuracy to full-voting methods while substantially reducing inference cost.
It matches Simple MV in average accuracy, both reaching 87.01\%, while reducing agent calls from 9.00 to 5.84 and token consumption from 7,381 to 4,123, corresponding to 35.1\% and 44.1\% reductions, respectively. 
Although Weighted MV achieves the highest accuracy of 87.33\%, it still requires all agents and 7,381 tokens per query. 
Overall, EMS offers a better accuracy-efficiency trade-off through reliability-aware ordering and certified early stopping.

\begin{table}
\centering
\small
\caption{Effect of different ordering and stopping strategies.
Random 5 and Random+ES use random agent orders without cold-start information.
For the other ordered variants, 5\% of the data is used for cold-start, and the reported efficiency metrics include this cold-start cost.
ES denotes early stopping.}
\label{tab:ablation_order_stop}
\begin{tabular}{l|c|c|c}
\toprule
{Methods} & {Avg. Acc.} & \text{Avg.\#Agents} & Avg.Tokens\\
\midrule
Full MV  & 87.01  & 9.00 &7,381 \\
\midrule
Random 5  & 85.59 & 5.00 & 4,065 \\ 
Fixed Top 5 & 86.40 & 5.27  & 4,374 \\
Random + ES  & 87.01 & 6.32 & 5,121 \\ 
Fixed + ES & 87.01 & 5.71  & 4,966 \\
EMS & 87.01 & 5.84&4,123  \\
\bottomrule
\end{tabular}
\end{table}

\subsection{Ablation Study}
In this section, we give a comprehensive analysis of the proposed method to verify its effectiveness.
\paragraph{Effect of Early Stopping.} 
The core insight of EMS is that an early stopping mechanism based on dynamic consensus checking can eliminate redundant computations without compromising reasoning quality. 
As shown in Table~\ref{tab:ablation_order_stop}, Full MV achieves an average accuracy of 87.01\%, but requires invoking all 9 agents and consumes 7,381 tokens per query.
Naively reducing the number of agents is not sufficient: Random 5 reduces the token cost to 4,065, but its accuracy drops to 85.59\%.
Similarly, Fixed Top 5 improves over Random 5, but still underperforms Full MV with an average accuracy of 86.40\%, while its fixed reliance on strong but verbose agents limits the token savings.
These results indicate that directly truncating the voting process can damage the reliability of multi-agent aggregation.
In contrast, early stopping preserves the accuracy of Full MV while reducing inference cost.
Random+ES reaches the same average accuracy as Full MV, while reducing Avg. \#Agents from 9.00 to 6.32 and Avg. Tokens from 7,381 to 5,121.
This shows consensus-based stopping is an effective safeguard: the system stops early only when the leading answer is sufficiently stable.

\paragraph{Effect of Agent Ordering Strategies.}
We further study whether a better agent order can improve the efficiency of early stopping.
Fixed+ES uses a fixed sequence of 9 agents learned from the 5\% cold start calibration set.
Compared with Random+ES, Fixed+ES maintains the same average accuracy of 87.01\%, while reducing Avg. \#Agents from 6.32 to 5.71 and Avg. Tokens from 5,121 to 4,966.
This suggests that even a globally fixed reliability-aware order can help the system reach a stable consensus earlier.
EMS further introduces Task-Conditioned Reliability Ordering.
It achieves the same average accuracy, while reducing Avg. Tokens to 4,123.
EMS reduces the average token consumption by 44.1\% compared with Full MV, and further saves 19.5\% and 17.0\% over Random+ES and Fixed+ES, respectively.
Although EMS invokes slightly more agents than Fixed+ES on average, it consumes fewer tokens, suggesting that  Task-Conditioned Reliability Ordering may select a more cost-effective agent sequence rather than merely minimizing the number of invoked agents.
Overall, EMS achieves the best accuracy-efficiency trade-off among the compared strategies.


\paragraph{Analysis of the Adaptive Incremental Voting.} 
To better understand how Adaptive Incremental Voting reduces redundant computation, we analyze the distribution of the number of invoked agents across benchmarks.
As shown in Figure~\ref{fig:1}, the Initial Voting stage with five agents already resolves a large portion of queries on relatively easier benchmarks.
For example, 90.1\% of GSM8K queries and 81.1\% of AQuA queries stop after the initial five agents.
In contrast, GPQA exhibits a much lower initial stopping rate of 33.3\%,  indicating that these more challenging tasks require additional agents to reach agreement. 
After the initial stage, EMS incrementally invokes extra agents only when the current votes are insufficient to determine the final winner.
For most benchmarks, only a small fraction of queries require all 9 agents.
This confirms that the certified stopping rule can adapt the voting depth to task difficulty: simple queries are resolved early, while difficult queries are allowed to collect more evidence.


\begin{table}[t]
\centering
\small
\setlength{\tabcolsep}{6pt}
\caption{
Early-stopping error analysis.
\textbf{ES}: early-stop rate; \textbf{TP}: correct early stops; \textbf{FP}: incorrect early stops.
\textbf{H-FP}: harmful false-positive stops corrected by FullMV.
\textbf{Shared Err.}: errors shared by early stopping and FullMV.
}
\label{tab:early_stop_error}
\begin{tabular}{ccc}
\toprule
Metric & Easy (AQuA) & Hard (GPQA) \\
\midrule
$n$ & 254 & 198 \\
ES & 87.80\% & 76.77\%  \\
TP & 81.89\% & 58.08\% \\
FP & 5.91\% (15) & 18.69\% (37)  \\
H-FP & 0.39\% (1) & 3.54\% (7)  \\
Shared Err. & 5.51\% (14) & 15.15\% (30) \\
\bottomrule
\end{tabular}
\end{table}

\paragraph{Early-Stopping Error Analysis.}
We further analyze whether early stopping introduces additional errors compared with Full MV.
Table~\ref{tab:early_stop_error} reports the early-stop rate and its error decomposition on an easy benchmark, AQuA, and a hard benchmark, GPQA.
On AQuA, EMS stops early on 87.80\% of the queries, among which 81.89\% are correct early stops.
The false-positive stop rate is only 5.91\%, and only 0.39\% of all queries are harmful false positives, i.e., cases where early stopping produces an incorrect answer that would have been corrected by Full MV.On GPQA, the task is substantially harder.
EMS still stops early on 76.77\% of the queries, but the false-positive stop rate increases to 18.69\%.
However, most of these errors are shared with Full MV, while only 3.54\% are harmful false positives.
This suggests that many early-stopping errors come from the intrinsic failure of the full voting pool rather than premature termination itself.

\begin{table}[t]
\centering
\small
\setlength{\tabcolsep}{6pt}
\caption{Performance scaling with respect to the maximum agent pool size $N$ on {GPQA}.}
\label{tab:scaling_gpqa}
\begin{tabular}{c|ccc}
\toprule
$N$ & Acc. & Avg.\#Agents & Avg.Tokens \\
\midrule
5   & 67.17 & 3.98 & 7,073  \\
7   & 68.18 & 5.43 & 8,725  \\
9   & {70.20} & 6.84 & 9,432  \\
11  & 71.21 & 8.17 & 10,547 \\
13  &  \textbf{71.72} & 9.56 & 12,250 \\
\bottomrule
\end{tabular}
\end{table}

\paragraph{Sensitivity Analysis of Agent Pool Size $N$.}
To investigate the scalability and upper-bound performance of EMS, we evaluate its behavior under different agent pool sizes $N \in \{5, 7, 9, 11, 13\}$ on the GPQA. The results are summarized in Table~\ref{tab:scaling_gpqa}. 
As the candidate pool size $N$ increases, we observe a consistent improvement in reasoning accuracy, accompanied by a gradual increase in the average number of agents participating in the voting process. 
However, the marginal benefit of enlarging the agent pool diminishes as $N$ grows. 
On the GPQA dataset, increasing $N$ from $9$ to $13$ requires invoking an additional 2.72 agents on average and increases token cost by 29.9\%, yet gains only a modest accuracy improvement of 1.52\%. 
This observation suggests that continually increasing the number of agents provides limited returns and may not be an efficient strategy for multi-agent voting. 
Furthermore, we observe that as $N$ increases, the average agent call rate of EMS remains relatively stable, 
indicating that the early-stopping mechanism effectively controls inference cost growth even when the agent pool becomes larger.

\section{Conclusion}
We study the efficiency of majority voting in multi-agent reasoning. 
Although majority voting provides a simple and effective way to aggregate diverse agent responses, traditional voting requires all agents to complete inference before aggregation resulting in redundant computation. 
We propose Efficient Majority-then-Stopping (EMS), which formulates efficient voting as a reliability-aware agent scheduling problem.
EMS orders agents by task-conditioned reliability, incrementally collects votes, terminates once the leading answer can no longer be overturned, and updates the
reliability history of invoked agents according to their consensus with the final consensus.
Experiments across five benchmarks demonstrate that EMS maintains the accuracy of majority voting while reducing the average number of invoked agents and token consumption. 


\newpage
\section*{Limitations}
Despite the effectiveness of EMS, our work has two key limitations. 
(1) The evaluation is merely conducted on the mathematical reasoning and general knowledge; more challenging and field tasks can be applied to verify the generalization of the proposed method.
(2) The essence of this algorithm is to improve efficiency by reducing the votes of unnecessary agents. However, based on human voting experience, the choice of voting agents is actually a very complex issue, which is also a key bottleneck.
(3) The efficiency of EMS depends on the quality of agent ordering, designing more expressive scheduling strategies remains an important direction for future work.

\section*{Ethical Considerations}
The proposed Efficient Majority-then-Stopping is an algorithm for general reasoning tasks, and the algorithm itself does not have any ethical risks. 

\bibliography{main}
\newpage
\appendix 
\section{Experimental Details}

\label{sec:appendix_datasets}

Following prior work~\cite{DBLP:conf/acl/WangW00Z0025}, we use the same dataset settings and evaluate our method on five benchmarks covering mathematical reasoning and commonsense knowledge. Table~\ref{tab:dataset_stats} summarizes the statistics and evaluation metrics for each dataset. We set the temperature $T=0$, Top-$p=1.0$, and random seed 42. The same answer extraction and correctness checking rules are applied to all methods.

\begin{table}[h]
    \centering
    \small 
    \setlength{\tabcolsep}{4pt}
    \caption{\textbf{Overview of Datasets.}}
    \label{tab:dataset_stats}
    \begin{tabular}{llc}
    \toprule
    \textbf{Dataset} & \textbf{Domain} & \textbf{Test Size} \\
    \midrule
    \multicolumn{3}{l}{\textit{Mathematical Reasoning}} \\
    GSM8K & Grade School Math & 1,319  \\
    AQuA & Algebra Problems & 254  \\
    \midrule
    \multicolumn{3}{l}{\textit{Commonsense \& General Knowledge}} \\
    MMLU & Academic Subjects & 1,474  \\
    CommonsenseQA & General Sense & 1,221  \\
    GPQA & Graduate Science & 198  \\
    \bottomrule
    \end{tabular}
    
\end{table}



\section{Additional Results}

\begin{table}
\centering
\small
\caption{Effect of different cold-start ratios.}
\label{tab:cold_start_analysis}
\begin{tabular}{l|c|c|c}
\toprule
{Setting} & {Avg. Acc.} & \text{Avg.\#Agents} & Avg.Tokens\\
\midrule
EMS-CS-${5\%}$  & 87.01 & 5.84 & 4,123 \\ 
EMS-CS-${10\%}$ & 87.01 & 6.01 & 4,455  \\
EMS-Onlie & 87.01 & 5.87 & 4,283  \\
\bottomrule
\end{tabular}
\end{table}
\subsection{Effect of Cold-Start Ratios}
As shown in Table~\ref{tab:cold_start_analysis}, we analyze the effect of the cold-start ratio in EMS, where EMS-CS-${5\%}$ and EMS-CS-${10\%}$ use 5\% and 10\% of the evaluation data to initialize the reliability estimates, respectively.  
For EMS-CS-${5\%}$ and EMS-CS-${10\%}$, the reported efficiency metrics are computed over the entire evaluation set and include the additional cost introduced by the cold-start stage, where all agents are invoked to initialize the reliability estimates. 

EMS-CS-${5\%}$ achieves the lowest token consumption and a slightly smaller number of invoked agents than EMS-CS-${10\%}$ and EMS-Online. 
Increasing the cold-start ratio to 10\% does not improve accuracy, but introduces more full-agent invocations during initialization, leading to higher average cost. 
The online variant avoids explicit cold-start data, but its early reliability estimates are less stable, resulting in slightly higher token consumption than EMS-CS-${5\%}$
Therefore, we use EMS-CS-${5\%}$ as the default setting, as it provides the best accuracy-efficiency trade-off under end-to-end cost accounting.

\subsection{Runtime Analysis}
\label{sec:appendix_runtime}

We further analyze the end-to-end runtime on AQUA. 
Full MV invokes all \(N=9\) agents in parallel and takes the maximum latency, while Random+ES and EMS invoke the initial batch of \(\tau=5\) agents in parallel and perform subsequent escalation sequentially.

\begin{table}[t]
\centering
\small
\caption{Runtime analysis on AQUA. LLM latency is recomputed according to the actual scheduling protocol. Computer overhead includes ACM scoring, stopping verification, and reliability updating.}
\label{tab:runtime_analysis}
\begin{tabular}{l|c|c|c}
\toprule
{Method} & \makecell{LLM\\Latency} & \makecell{Computer\\Overhead} & \makecell{Total\\Latency} \\
\midrule
Full MV      & 26.306s/q & -  & 26.306s/q \\
Random+ES    & 19.022s/q & - & 19.031s/q \\
EMS     & 14.146s/q & 16.007ms/q & 14.162s/q \\
\bottomrule
\end{tabular}
\end{table}

Although EMS may perform sequential escalation after the initial parallel batch, its additional computational overhead is negligible, requiring only 16.007 ms per question. 
However, it does not require all agents to be invoked for most questions. 
Therefore, the runtime gain comes from reducing the number of LLM calls on the effective execution path.
\end{document}